\documentclass{article}

\usepackage[preprint]{neurips_2025}

\usepackage[utf8]{inputenc}
\usepackage{bm}  

\usepackage[table]{xcolor}
\usepackage{booktabs}
\usepackage{array}
\usepackage{tcolorbox}

\definecolor{rowE2E}{HTML}{F2E9DD}   
\definecolor{muted}{HTML}{7D6951}    
 
\definecolor{deltaGood}{HTML}{2F7A4F}     
\definecolor{deltaBad}{HTML}{B05B3B}      
\definecolor{muted}{HTML}{7D6951}
\definecolor{groupRow}{HTML}{F4EEE6}      

\newcommand{\dgood}[1]{\textcolor{deltaGood}{#1}}
\newcommand{\dbad}[1]{\textcolor{deltaBad}{#1}}
\usepackage[T1]{fontenc}
\usepackage{hyperref}
\usepackage{url}
\usepackage{booktabs}
\usepackage{amsfonts}
\usepackage{amsmath,amssymb}
\usepackage{nicefrac}
\usepackage{microtype}
\usepackage{xcolor}
\usepackage{graphicx}
\usepackage{subcaption}
\usepackage{algorithm}
\usepackage{algorithmic}
\usepackage{cleveref}
\usepackage{bm}
\usepackage{bbm}
\usepackage{mathtools}
\usepackage{multirow}
\usepackage{enumitem}
\usepackage{wrapfig}
\usepackage{tabularx}
\usepackage{longtable}
\usepackage{placeins}

\newcommand{\xb}{\bm{x}}
\newcommand{\zb}{\bm{z}}

\newcommand{\thetab}{\bm{\theta}}

\title{ReflectDrive-2: Reinforcement-Learning-Aligned Self-Editing for Discrete Diffusion Driving}

\author{
\makebox[\textwidth][c]{%
\textbf{Huimin Wang\textsuperscript{*‡}},
Yue Wang\textsuperscript{*},
\textbf{Bihao Cui\textsuperscript{†}}
}\\
\\
\makebox[\textwidth][c]{%
\textbf{
Pengxiang Li,
Ben Lu,
Mingqian Wang,
Tong Wang,
Chuan Tang,
Teng Zhang,
Kun Zhan
}}\\
\\
\makebox[\textwidth][c]{%
LiAuto
}
}

\begin{document}

\maketitle
\begingroup
\renewcommand\thefootnote{*}
\footnotetext{Equal Contribution.}
\endgroup

\begingroup
\renewcommand\thefootnote{†}
\footnotetext{Project Lead.}
\endgroup

\begingroup
\renewcommand\thefootnote{‡}
\footnotetext{\texttt{wanghuimin1@lixiang.com}}  
\endgroup

\begin{abstract}
We introduce ReflectDrive-2, a masked discrete diffusion planner with a separate action expert for autonomous driving that represents plans as discrete trajectory tokens and generates them through parallel masked decoding. This discrete token space enables in-place trajectory revision: AutoEdit rewrites selected tokens using the same model, without requiring an auxiliary refinement network. To train this capability, we use a two-stage procedure. First, we construct structure-aware perturbations of expert trajectories along longitudinal progress and lateral heading directions and supervise the model to recover the original expert trajectory. We then fine-tune the full decision--draft--reflect rollout with reinforcement learning (RL), assigning terminal driving reward to the final post-edit trajectory and propagating policy-gradient credit through full-rollout transitions. Full-rollout RL proves crucial for coupling drafting and editing: under supervised training alone, inference-time AutoEdit improves PDMS by at most $0.3$, whereas RL increases its gain to $1.9$. We also co-design an efficient reflective decoding stack for the decision--draft--reflect pipeline, combining shared-prefix KV reuse, Alternating Step Decode, and fused on-device unmasking. On NAVSIM, ReflectDrive-2 achieves $91.0$ PDMS with camera-only input and $94.8$ PDMS in a best-of-6 oracle setting, while running at $30.2$ ms average latency on NVIDIA Thor.
\end{abstract}

\section{Introduction}
\label{sec:intro}
\begin{figure}[t]
  \centering
  \includegraphics[width=\linewidth]{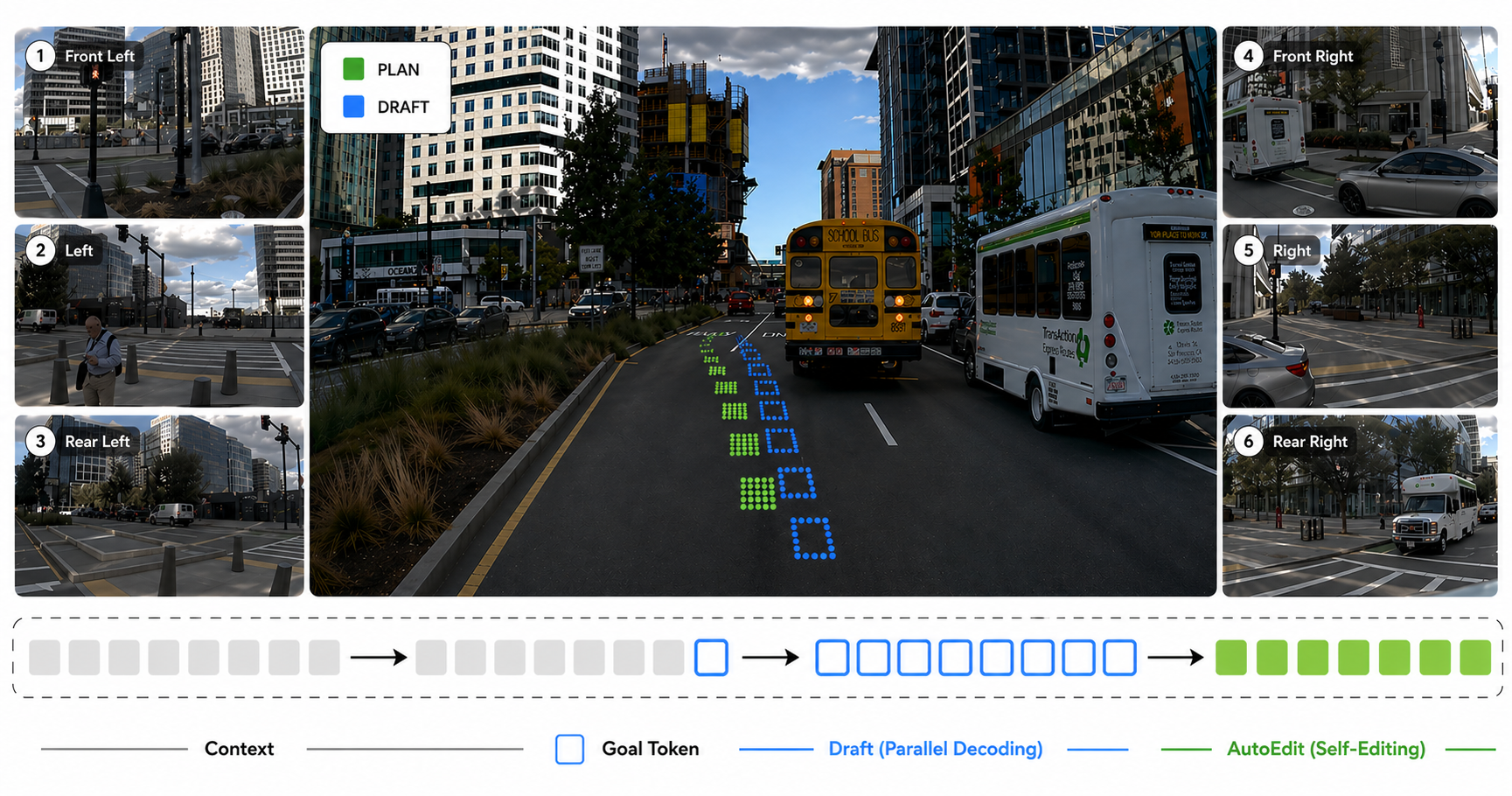}
  \caption{\textbf{ReflectDrive-2 plans through a decision–draft–reflect process in a shared discrete token space.} Conditioned on surround-view cameras, navigation, and ego-state, a goal-point posterior commits a single Goal Token, masked discrete diffusion parallel-decodes a trajectory Draft (blue), and AutoEdit rewrites tokens in place to yield the final Plan (green).}
  \label{fig:teaser}
\end{figure}
Planning errors in imitation-learned driving policies are rarely random. They concentrate along two common axes: longitudinal speed misjudgment (overshoot, under-progress, late braking) and lateral heading drift (lane deviation, clipped turns, drivable-area violations). These are the directions along which imitation learning from expert demonstrations accumulates covariate shift~\citep{bansal2018chauffeurnet,codevilla2018end}, and they are the directions along which an in-place correction mechanism could act. A planning representation that supports structured in-place revision is therefore well-matched to the error structure of the problem.
Classical modular stacks~\citep{fan2018apollo,kato2018autoware} and end-to-end planners~\citep{bojarski2016end,hu2023uniad,chitta2022transfuser,hu2022stp3,jiang2023vad} commit to a single trajectory; autoregressive vision-language-action (VLA) planners~\citep{kim2024openvla,tian2024drivevlm,sima2024drivelm} inherit sequential decoding and revise emitted tokens only by re-rolling the full sequence; continuous diffusion planners~\citep{janner2022diffuser,chi2023diffusionpolicy,li2024diffusiondrive,zhou2024goalflow} parallelize generation but reverse a Gaussian corruption process rather than the structured failure modes of a trained driver. Masked discrete diffusion~\citep{austin2021d3pm,nie2025llada,song2025seed,bie2025llada21} admits such revision natively: any subset of trajectory tokens can be re-masked and rewritten by the same model, conditioned on the rest, without an auxiliary network or a separate inference mode.

Simply adding a self-editing step on top of a trained drafter, however, yields little. The drafter has no incentive to emit drafts that the editor can improve, and the editor receives no signal indicating which rewrites pay off in closed-loop behavior. Under supervised training alone, the self-editing capability exists in the weights but the two stages are decoupled: the drafter optimizes its own token-level loss, and the editor optimizes a separate correction loss. Neither stage is aware of the other's effect on the final driving outcome.
Reinforcement learning (RL) over the full draft-and-edit rollout closes this gap. When a single terminal reward assigns policy-gradient credit to both drafting and editing transitions, the two phases become coupled. The drafter learns to emit revisable drafts -- token distributions whose post-edit trajectory scores higher than the pre-edit one -- and the editor learns corrections that move the draft toward the closed-loop reward rather than only reducing token-level uncertainty. Self-correction is no longer a post-hoc add-on; it becomes part of the optimized policy rollout.

We call the resulting system \textit{ReflectDrive-2}, a reflective masked-diffusion VLA planner, and its self-editing mechanism \textbf{AutoEdit}. ReflectDrive-2's inputs are panoramic cameras, route/navigation instruction tokens, and ego state; its outputs are discrete trajectory tokens whose final waypoint tokens anchor a behavior hypothesis, and whose remaining trajectory tokens realize the 4-second plan. Each goal point represents a candidate behavioral hypothesis, such as lane keeping, yielding, overtaking, or changing lanes, and is selected from the predicted goal posterior using top-$k$ sampling with non-maximum suppression. AutoEdit is pretrained against structure-aware perturbations spanning the longitudinal and lateral failure axes above, and then co-trained with the drafter through RL over the joint rollout. Vision and natural-language instructions serve as joint conditioning inputs to a shared backbone that denoises discrete action tokens, and drafting together with AutoEdit constitutes a unified policy loop optimized using a single reward signal.

The reflective structure also shapes the runtime. The inference path (context prefill, goal proposal, multi-batch drafting, AutoEdit) admits a reflection-aware stack with shared-prefix KV cache reuse across the decision--draft--reflect phases, Alternating Step Decode (ASD) that reuses AutoEdit across frames as a temporal refiner, and a fused on-device unmasking kernel. On NAVSIM~\citep{dauner2024navsim}, ReflectDrive-2 reaches $91.0$ PDMS camera-only, and $94.8$ PDMS under best-of-$6$ oracle selection; on NVIDIA Thor the stack averages $30.2$ ms per frame.

To summarize, our main contributions are as follows:
\begin{itemize}
    \item \textbf{Goal-conditioned masked-diffusion planning.}
    We propose \textit{ReflectDrive-2}, a driving VLA that plans through a \textit{decision--draft--reflect} process. A goal-point posterior exposes behavior-level hypotheses; masked discrete diffusion drafts editable trajectories for each hypothesis; and AutoEdit rewrites drafts in the same token space. On NAVSIM, ReflectDrive-2 achieves $91.0$ PDMS with camera-only input, and $94.8$ PDMS under best-of-$6$ oracle selection.
    \item \textbf{Reward-coupled AutoEdit.}
    We introduce \textbf{AutoEdit}, a self-correction mechanism trained with structure-aware perturbations that match the longitudinal and lateral failure axes of imitation-learned driving. By applying RL over the full draft-and-edit rollout, the reward signal co-adapts drafter and editor, substantially amplifying the effectiveness of inference-time AutoEdit.
    \item \textbf{Efficient reflective decoding.}
    We co-design a runtime stack that exploits the decision--draft--reflect structure: shared-prefix KV cache, ASD reinterpreted as temporal AutoEdit, and fused CUDA unmasking, achieving $30.2$ ms average latency on NVIDIA Thor with near-lossless planning quality.
\end{itemize}
\section{Related Work}
\label{sec:related}

\subsection{End-to-End and VLA Planning}
\label{sec:related:e2e}

End-to-end planners map sensors to trajectories without inter-module error propagation~\citep{chitta2022transfuser,hu2023uniad,jiang2023vad,hu2022stp3}; SMART~\citep{wu2024smart} tokenizes multi-agent trajectories for autoregressive next-token prediction. VLA planners~\citep{jin2024drivevla,zhou2024autovla,zhang2024recogdrive,kim2024openvla,tian2024drivevlm,sima2024drivelm} inherit language priors but decode token-by-token, so latency scales with trajectory length and any correction requires a second sequential rollout. Continuous diffusion planners~\citep{janner2022diffuser,chi2023diffusionpolicy,li2024diffusiondrive,zhou2024goalflow,zheng2026hdp} generate in parallel but require $\geq 20$ denoising steps, and guided variants~\citep{zhong2023ctgpp,jiang2023motiondiffuser} compound cost through per-step gradient propagation. ReflectDrive-2 replaces both paradigms with masked discrete diffusion: parallel unmasking reaches a full trajectory in a few rounds, and token-level editing is native rather than a second-stage add-on. These baselines do not naturally couple in-place editing with the same policy rollout and reward signal -- the property that our approach builds on.

\subsection{Discrete Diffusion and Token-Space Editing}
\label{sec:related:discrete}

Discrete diffusion provides a natural generative framework for categorical state spaces. 
D3PM~\citep{austin2021d3pm} extends diffusion modeling to discrete variables, and 
MaskGIT~\citep{chang2022maskgit} shows that masked-token prediction can support 
parallel generation through confidence-based unmasking. This line has recently scaled 
to language modeling: LLaDA~\citep{nie2025llada} and Seed Diffusion~\citep{song2025seed} 
train large masked-diffusion language models, while MDLM~\citep{lou2024mdlm}, 
SEDD~\citep{lou2024sedd}, Block Diffusion~\citep{arriola2025blockdiffusion}, and 
Fast-dLLM~\citep{wu2025fastdllm} improve the formulation or serving efficiency of 
discrete diffusion models. LLaDA 2.0/2.1~\citep{bie2025llada20,bie2025llada21} further 
scale this paradigm and introduce Token-to-Token (T2T) editing, where low-confidence 
tokens are regenerated during decoding.

The ability to re-mask and regenerate arbitrary token subsets makes discrete diffusion 
especially suitable for editable planning. However, most existing token-editing 
mechanisms are either decoding-time heuristics or independently trained refinement 
stages. LLaDA 2.1 T2T~\citep{bie2025llada21}, for example, revises tokens according 
to model confidence, but the model is not explicitly trained on the structured errors 
that arise in downstream control. In contrast, AutoEdit is supervised with trajectory 
perturbations aligned with common driving failure modes, including longitudinal 
progress errors and lateral heading deviations. The editor therefore observes the 
types of failures it is expected to correct during training, rather than relying only 
on uncertainty estimates at inference time.

Recent work has also explored refinement in embodied or multimodal diffusion models. 
DriveFine~\citep{dang2026drivefine} is the closest prior work, introducing a 
refinement-augmented masked-diffusion driving VLA. Its refiner, however, is trained 
and optimized separately from the drafter. ReflectDrive-2 instead treats drafting and 
editing as a single composed rollout: the terminal driving reward is assigned to the 
post-edit trajectory, and policy-gradient credit is applied to token transitions from 
both stages. This joint credit assignment allows the drafter and editor to co-adapt 
under the same closed-loop objective. Similarly, ``From denoising to refining'' 
\citep{ji2025denoising} studies corrective refinement for vision--language diffusion 
models, but focuses on multimodal understanding rather than closed-loop control and 
does not couple the refiner to a driving reward. LLaDA-VLA~\citep{wen2025lladavla} 
applies discrete diffusion to robot control, while ReflectDrive-2 focuses on 
token-space editing for autonomous driving and optimizes the draft--edit process 
through a shared rollout reward.

\subsection{Reinforcement Learning for Diffusion Policies}
\label{sec:related:rl}

DDPO~\citep{black2024ddpo} and DPPO~\citep{fan2024dppo} apply policy gradients to continuous diffusion by treating denoising as a multi-step MDP, which requires reparameterization in continuous state spaces. For discrete diffusion, d1~\citep{zhao2025d1} uses GRPO-style RL but ignores multi-step structure; d2~\citep{wang2025d2} recovers it with step-aware gradients and group-relative advantage; SPG~\citep{wang2025spg} derives tighter ELBO/EUBO bounds. In driving, HDP~\citep{zheng2026hdp} and DriveFine~\citep{dang2026drivefine} adopt RL post-training on diffusion planners. These methods each optimize a \emph{single-pass} rollout: drafting alone, or refining alone. ReflectDrive-2's RL objective is applied to a \emph{composed} rollout, $\text{draft}\!\to\!\text{AutoEdit}$, so the terminal reward credits both stages jointly. Simply increasing the number of diffusion steps does not expose a semantically distinct edit operator to receive reward credit; our composed rollout contains a reflection phase that shares the reward with drafting. \Cref{sec:method:rl} formalizes the distinction and \Cref{tab:autoedit_gain} isolates the substantial amplification of the editor's gain that results.

\section{Preliminaries}
\label{sec:prelim}

\subsection{Problem Setting}
\label{sec:prelim:problem}

At time step $t$, the ego vehicle receives an observation $\bm{o}_t = (\bm{v}_t, \bm{\ell}_t, \bm{s}_t)$ with three channels: panoramic visual tokens $\bm{v}_t$ from left-front, front, and right-front cameras over two temporal frames; a \emph{navigational instruction} channel $\bm{\ell}_t$ carrying route-level commands and maneuver hints (keep lane, turn left at intersection, proceed straight) as linguistic tokens consumed by the same backbone that models action tokens; and an ego-state channel $\bm{s}_t$ with kinematic tokens (velocity, acceleration, yaw rate). The instruction channel is the ``L'' of our VLA: it conditions drafting on intent, not just on scene. The objective is to generate a future trajectory $\bm{\tau}=\{(x_k,y_k)\}_{k=1}^{K}$ that is safe, comfortable, rule-compliant, and consistent with $\bm{\ell}_t$. Heading is derived from consecutive waypoints when required by downstream metrics.

\subsection{Masked Discrete Diffusion}
\label{sec:prelim:discrete}

\paragraph{Forward and reverse process.}
We represent the future ego trajectory as a sequence of Bird's-Eye-View (BEV) coordinate tokens, denoted by $\mathbf{x}_0$. Following masked discrete diffusion~\citep{austin2021d3pm,nie2025llada}, the forward process corrupts $\mathbf{x}_0$ by independently replacing each token with \texttt{[MASK]} at probability $t \in [0,1]$, yielding a partially masked sequence $\mathbf{x}_t$. A bidirectional Transformer $p_{\thetab}$ reverses this process by predicting the original tokens from $\mathbf{x}_t$ conditioned on multimodal context $\mathbf{c}=(\bm{v}_t,\bm{\ell}_t,\bm{s}_t)$. Prior masked-diffusion language models typically optimize a $1/t$-weighted cross-entropy on masked positions only~\citep{nie2025llada}; we supervise \emph{all} positions:
\begin{equation}
    \mathcal{L}_{\text{DLM}}(\theta)=
    -\mathbb{E}_{\mathbf{x}_0,\, t}
    \left[
    \frac{1}{L}\sum_{i=1}^{L}
    \log p_\theta(x_0^i \mid \mathbf{x}_t, \mathbf{c})
    \right].
    \label{eq:dlm_loss}
\end{equation}
Empirically the all-position objective yields more stable optimization and coherent drafts. At inference time, generation begins from a fully masked sequence and proceeds through a small number of parallel denoising steps.

\paragraph{Selective re-generation.}
Masked diffusion admits arbitrary in-place rewriting: for any edit mask $\bm{e}\in\{0,1\}^L$, the partial sequence $\tilde{\xb}=\bm{e}\odot\texttt{[MASK]}+(1-\bm{e})\odot\xb_0$ is denoised from effective time $t^{*}\!\approx\!|\bm{e}|/L$. LLaDA 2.1~\citep{bie2025llada21} extends this idea through Token-to-Token (T2T) editing, which also revises low-confidence tokens at decoding time. Our AutoEdit framework inherits this interface but shifts the editor from decoding-time heuristic to trained operator (\Cref{sec:method:autoedit}) and couples it to the drafter through a shared RL reward (\Cref{sec:method:rl}).

\subsection{KV Caching for Efficient Inference}
\label{sec:prelim:kvcache}
Standard masked diffusion uses bidirectional attention, so vanilla KV caching fails: KV entries must be recomputed at every denoising step because masked tokens change~\citep{nie2025llada}. Block Diffusion~\citep{arriola2025blockdiffusion} partitions the sequence into blocks, running diffusion within a block and generating blocks autoregressively for cache reuse on completed blocks. LLaDA 2.1~\citep{bie2025llada21} generalizes to block-wise causal attention, and LLaDA 2.0~\citep{bie2025llada20} adds serving-level optimizations such as variable-length batching and prefix caching in its dInfer engine. We adopt causal attention over the scene-context prompt and block-wise attention over trajectory tokens, which permits KV reuse for the prompt while preserving bidirectional diffusion within the trajectory block (\Cref{sec:deployment}).

\subsection{Reinforcement Learning Fine-Tuning}
\label{sec:prelim:rl}

Supervised training imitates the data distribution but does not optimize driving objectives directly. We cast trajectory generation as a Markov decision process and fine-tune with reinforcement learning so the policy is aligned with a closed-loop reward. Following~\citet{wang2025d2}, the objective is $J(\theta) = \mathbb{E}_{\tau\sim\pi_\theta(\cdot\mid o)}[R(\tau)]$, optimized with group-relative advantage over $G$ sampled trajectories and a discrete-diffusion policy gradient:
\begin{equation}
\begin{aligned}
\mathcal{L}(\theta) = -\mathbb{E}_{\tau_{1:G} \sim \pi_{\theta_{\mathrm{old}}}(\cdot|o)} \bigg[ &\frac{1}{G} \sum_{g=1}^{G} \frac{1}{L} \sum_{s=1}^{S} \sum_{p=1}^{L} \mathbf{1}_{\{x_{g,p}^{s+1} \neq x_{g,p}^{s}\}} \cdot \\
&\min\big( r_{g,p}^{s} A_g, \mathrm{clip}(r_{g,p}^{s}, 1-\epsilon, 1+\epsilon) A_g \big) \bigg] + \lambda_{\mathrm{KL}} D_{\mathrm{KL}}\left( \pi_\theta \,\|\, \pi_{\mathrm{ref}} \right),
\end{aligned}
\label{eq:total_loss}
\end{equation}
where $S$ is the total number of generation steps, $A_g=R(\tau_g)-\tfrac{1}{G}\sum_j R(\tau_j)$, and $r_{g,p}^{s}=\pi_\theta(x_{g,p}^s\mid\mathbf{x}_{g,s+1},o)/\pi_{\theta_{\mathrm{old}}}(x_{g,p}^s\mid\mathbf{x}_{g,s+1},o)$. The indicator $\mathbf{1}_{\{x_{g,p}^{s+1}\neq x_{g,p}^{s}\}}$ restricts credit to tokens that are actually updated at step $s$. In \Cref{sec:method:rl} we instantiate $S=S_{\mathrm{draft}}+S_{\mathrm{edit}}$, so the same reward credits token transitions from drafting and AutoEdit jointly -- the methodological centerpiece of this paper.

\section{Method}
\label{sec:method}

\subsection{ReflectDrive-2 Overview}
\label{sec:method:overview}

ReflectDrive-2 formulates autonomous driving planning as goal proposal, masked trajectory drafting, and token-space trajectory correction within a unified discrete representation. 
Given multimodal driving context $\mathbf{c}=(\bm{v}_t,\bm{\ell}_t,\bm{s}_t)$, where $\bm{v}_t$ denotes visual tokens, $\bm{\ell}_t$ denotes route-instruction tokens, and $\bm{s}_t$ denotes ego-state tokens, the model first predicts a set of goal-point hypotheses. 
Each goal is then used to condition a masked discrete-diffusion decoder that generates a trajectory in parallel over a small number of denoising rounds. 
After the initial draft is produced, AutoEdit reuses the same conditional token model to update selected trajectory tokens. 
The planner therefore performs generation and correction in the same action-token space, without introducing a separate refinement network.

The method has three coupled components. 
First, a goal-point posterior provides a compact decision layer over behavior-level hypotheses, such as different turning lines, yielding behavior, or passing around another agent. 
Second, goal-conditioned masked diffusion realizes each selected hypothesis as a full trajectory by filling discrete BEV coordinate tokens. 
Third, AutoEdit performs token-space correction by selectively rewriting parts of the drafted trajectory. 
The supervised stage trains both masked trajectory generation and structure-aware correction: standard random masking teaches the model to draft trajectories, while perturbation-based correction teaches it to recover clean trajectories from longitudinal and lateral planning errors. 
A constraint-aware field loss further regularizes the spatial distribution of predicted tokens against drivable-area geometry.

The reinforcement-learning stage optimizes the complete draft-and-edit rollout rather than the drafting stage alone. 
For each sampled candidate, the terminal driving reward is assigned to the final post-edit trajectory, and policy-gradient credit is applied to token transitions from both the drafting and AutoEdit phases. 
This coupling is central to ReflectDrive-2: AutoEdit is not treated as a post-processing heuristic, but as part of the policy rollout that is optimized under the same closed-loop objective as the drafter. 
The complete inference path can be summarized as
\begin{equation}
    \mathbf{c}
    \rightarrow
    \{g_m\}_{m=1}^{N_g}
    \rightarrow
    \{\xb_m^{(0)}\}_{m=1}^{N_g}
    \rightarrow
    \{\xb_m^{(K)}\}_{m=1}^{N_g},
    \label{eq:decision_draft_reflect}
\end{equation}
where $g_m$ is a sampled goal point, $\xb_m^{(0)}$ is the drafted trajectory conditioned on $g_m$, and $\xb_m^{(K)}$ is the trajectory after $K$ AutoEdit rounds. 
Vision tokens, route-instruction tokens, ego-state tokens, goal tokens, and trajectory tokens are processed by the same backbone, while diffusion denoising is applied to the action-token block. 
This shared token substrate allows trajectory drafting and editing to be trained and optimized as one action-generation process.

\begin{figure}[t]
  \centering
  \includegraphics[width=\linewidth]{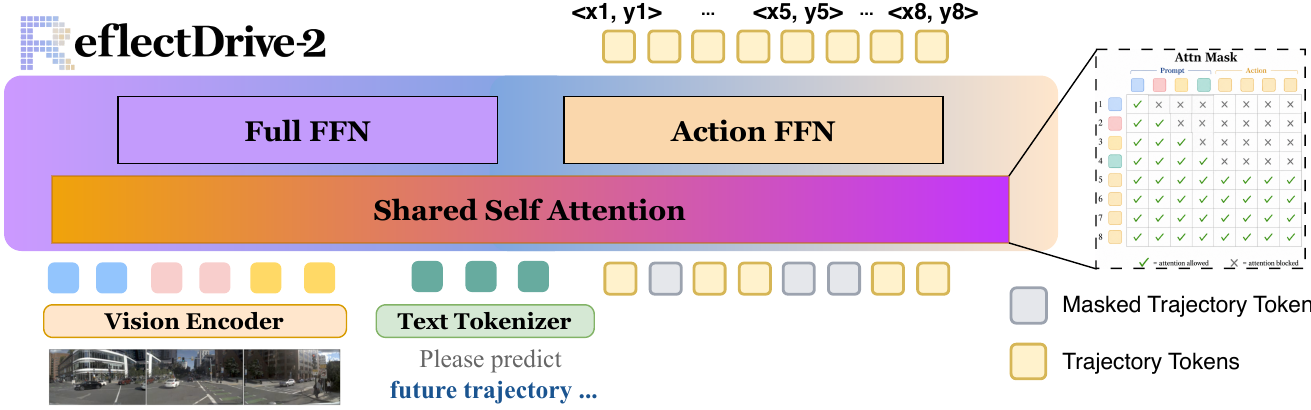}
  \caption{\textbf{ReflectDrive-2 architecture.} Vision, language, ego-state, and trajectory tokens share a single self-attention backbone, with a Full FFN over prompt tokens and a compact Action FFN over the action-token block. The attention pattern (right) is causal over the prompt, enabling shared-prefix KV reuse and block-wise bidirectional over action tokens.}
  \label{fig:architecture}
\end{figure}

\subsection{Goal-Conditioned Masked Trajectory Diffusion}
\label{sec:method:drafting}

\paragraph{Multimodal context encoding.}
Two temporally adjacent panoramic frames from the left-front, front, and right-front cameras are encoded by a ViT visual backbone and projected into the diffusion Transformer's token space. 
The resulting visual tokens are concatenated with route-instruction tokens $\bm{\ell}_t$ and ego-state tokens $\bm{s}_t$, and the concatenated sequence is processed by the shared backbone. 
Each Transformer block additionally contains an action-specific FFN and an action head, which specialize the model for trajectory-token prediction while retaining the shared backbone for scene and context modeling.

\paragraph{Goal-point prediction.}
Rather than committing to a unimodal endpoint prediction, ReflectDrive-2 predicts a goal-point posterior over discrete BEV coordinates. 
A goal point is represented as a discrete $(x,y)$ token pair and serves as a behavior-level hypothesis for the future plan. 
During training, the goal head is supervised by the expert endpoint. 
During inference, we sample candidate goals using top-$k$ sampling followed by non-maximum suppression (NMS) in BEV space. 
NMS removes duplicate endpoints while preserving spatially distinct alternatives, so different surviving goals can correspond to different maneuvers, such as lane keeping versus yielding, pass-left versus pass-right, or different feasible lines through a turn. 
Each selected goal conditions a separate masked-diffusion drafting branch.

\paragraph{Masked trajectory drafting.}
We represent the future ego trajectory over the benchmark planning horizon with $8$ waypoints. 
Each waypoint is discretized into one longitudinal and one lateral coordinate token, yielding a length-$L=16$ trajectory sequence
\begin{equation}
    \xb_0 = [x_1,y_1,\ldots,x_8,y_8],
\end{equation}
where the final coordinate pair $(x_8,y_8)$ corresponds to the selected goal. 
During supervised training, random positions are replaced by \texttt{[MASK]} and the model is trained with the all-position masked-diffusion objective in Eq.~\eqref{eq:dlm_loss}. 
At inference time, the selected goal tokens are fixed, the remaining trajectory tokens are initialized as \texttt{[MASK]}, and the model fills masked positions over a small number of parallel denoising rounds. 
At each round, the most confident predictions are committed. 
The generation cost is therefore determined by the number of denoising rounds rather than the number of trajectory tokens, and the same masked-token interface later enables selective trajectory rewriting.

\subsection{AutoEdit Trajectory Correction}
\label{sec:method:autoedit}

AutoEdit is a token-to-token trajectory editor operating in the same discrete action space as the
masked-diffusion drafter. 
Unlike masked trajectory drafting, AutoEdit does not convert selected trajectory tokens back to
\texttt{[MASK]}. 
Instead, it takes the current concrete trajectory-token sequence as input, predicts replacement tokens
at trajectory positions, and commits only the selected replacements. 
Thus, AutoEdit performs direct token-to-token rewriting rather than re-masking and re-denoising.

\paragraph{Structure-aware perturbations.}
Given a clean waypoint sequence $\zb_0=\{\zb_i\}_{i=1}^{N}$, we synthesize a perturbed trajectory
$\tilde{\zb}_0=\mathcal{T}(\zb_0)$ before tokenization. 
The perturbation operator $\mathcal{T}$ targets two common planning-error families: longitudinal
progress errors and lateral heading deviations.

\emph{Longitudinal progress perturbation.}
We rescale progress along the trajectory arc length:
\begin{equation}
    \tilde{\zb}_i = \mathrm{Interp}(\zb_0, \beta d_i),
    \qquad
    \beta \sim \mathcal{U}(\beta_{\min}, \beta_{\max}),
    \label{eq:longitudinal_perturb}
\end{equation}
where $d_i$ is the arc length at waypoint $\zb_i$. 
Values $\beta<1$ produce conservative under-progress, while $\beta>1$ produces overshoot or
insufficient deceleration.

\emph{Lateral heading perturbation.}
We rotate the trajectory in the ego frame:
\begin{equation}
    \tilde{\zb}_i = \mathbf{R}(\alpha)\zb_i,
    \qquad
    \mathbf{R}(\alpha)=
    \begin{bmatrix}
        \cos\alpha & -\sin\alpha\\
        \sin\alpha & \cos\alpha
    \end{bmatrix},
    \qquad
    \alpha \sim \mathcal{U}(-\alpha_{\max}, \alpha_{\max}).
    \label{eq:lateral_perturb}
\end{equation}
This produces coherent lateral deviation while preserving trajectory smoothness.

After tokenizing the perturbed trajectory into $\tilde{\xb}_0$, AutoEdit is trained to map the perturbed
token sequence directly back to the clean token sequence:
\begin{equation}
    q_\theta(\cdot \mid \tilde{\xb}_0, \mathbf{c})
    =
    \operatorname{softmax}
    \big(
      h_\theta(\tilde{\xb}_0, \mathbf{c})
    \big),
    \label{eq:autoedit_t2t_distribution}
\end{equation}
where $h_\theta$ is the shared conditional token model used by the planner. 
The structure-aware AutoEdit loss is
\begin{equation}
    \mathcal{L}_{\mathrm{SAP}}(\theta)
    =
    -\mathbb{E}_{\xb_0,\mathcal{T}}
    \left[
        \frac{1}{L}\sum_{i=1}^{L}
        \log q_\theta(x_0^i \mid \tilde{\xb}_0, \mathbf{c})
    \right].
    \label{eq:sap_loss}
\end{equation}
This objective teaches the model to directly translate perturbed trajectory tokens into clean
trajectory tokens, rather than to recover clean tokens from a newly masked sequence.

\paragraph{Inference-time AutoEdit.}
At test time, AutoEdit starts from a drafted trajectory $\xb^{(0)}$ and performs $K$ token-to-token
editing rounds. 
At round $k$, the model predicts a replacement-token sequence from the current trajectory tokens:
\begin{equation}
    \hat{\xb}^{(k+1)}
    =
    \operatorname{T2TEdit}_{\theta}
    \left(
        \xb^{(k)}, \mathbf{c}
    \right).
    \label{eq:autoedit_t2t_predict}
\end{equation}
We then compute a commit mask $\bm{m}^{(k)}\in\{0,1\}^{L}$, where $\bm{m}^{(k)}_i=1$ means that
the replacement token at position $i$ is committed. 
In the default setting, $\bm{m}^{(k)}$ selects low-confidence non-goal trajectory tokens, while the
goal tokens remain fixed as the behavior anchor. 
The update is
\begin{equation}
    \xb^{(k+1)}
    =
    \bm{m}^{(k)}\odot \hat{\xb}^{(k+1)}
    +
    \left(1-\bm{m}^{(k)}\right)\odot \xb^{(k)}.
    \label{eq:autoedit_t2t_update}
\end{equation}
Importantly, $\bm{m}^{(k)}$ is a commit mask, not a re-masking mask. the input to AutoEdit remains
the concrete token sequence $\xb^{(k)}$, and no selected token is converted back to \texttt{[MASK]}.
AutoEdit is therefore a direct token-to-token trajectory editor implemented by the same conditional
token model as the drafter, without an auxiliary refinement network or a hand-designed smoothing
module.

\begin{figure}[t]
  \centering
  \includegraphics[width=\linewidth]{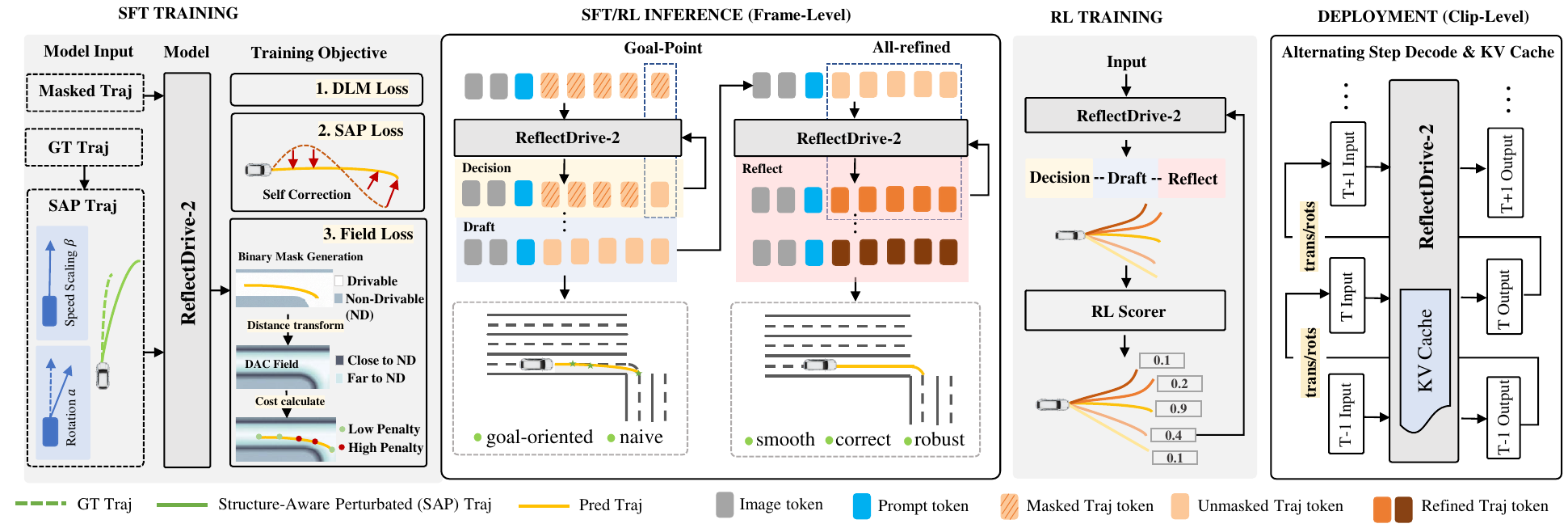}
  \caption{\textbf{Training, inference, and deployment of ReflectDrive-2.}
\textbf{(a)} SFT with three objectives: masked-token loss $\mathcal{L}_{\text{DLM}}$, structure-aware perturbation loss $\mathcal{L}_{\text{SAP}}$ over longitudinal scaling $\beta$ and lateral rotation $\alpha$, and drivable-area field loss $\mathcal{L}_{\text{field}}$.
\textbf{(b)} Frame-level \textit{decision--draft--reflect} inference: commit a goal token, parallel-decode the draft, then rewrite tokens in place via AutoEdit.
\textbf{(c)} RL fine-tuning scores full draft-and-edit rollouts with a closed-loop reward, propagating group-relative credit through both phases.
\textbf{(d)} Clip-level deployment alternates full-step and lite-step frames with shared-prefix KV reuse; rigid-body transforms carry the previous plan into the current ego frame.}
  \label{fig:training_inference}
\end{figure}

\subsection{Constraint-Aware Supervised Objectives}
\label{sec:method:sup}

\paragraph{Drivable-area field loss.}
The masked-diffusion loss $\mathcal{L}_{\mathrm{DLM}}$ and the AutoEdit correction loss $\mathcal{L}_{\mathrm{SAP}}$ optimize token-level prediction, but they do not explicitly encode drivable-area geometry. 
We add a field-based spatial penalty over the waypoint distribution induced by the coordinate-token logits. 
Let $p_{xy}^{(t)}\in\mathbb{R}^{H\times W}$ denote the spatial distribution at waypoint $t$, obtained from the marginal coordinate distributions as
\begin{equation}
    p_{xy}^{(t)}[i,j] = p_x^{(t)}[i]\,p_y^{(t)}[j].
\end{equation}
Given a BEV cost field $\mathcal{C}\in\mathbb{R}_{\ge 0}^{H\times W}$, we penalize probability mass assigned to high-cost cells using a field-weighted log barrier:
\begin{equation}
    \mathcal{L}_{\mathrm{field}}
    =
    \sum_{t=1}^{T}\sum_{i,j}
    -\log\!\left(1-p_{xy}^{(t)}[i,j]\right)\mathcal{C}[i,j].
    \label{eq:field_loss}
\end{equation}
The logarithmic factor gives larger gradients when the model assigns high confidence to high-cost regions.

In our implementation, $\mathcal{C}$ is instantiated as a drivable-area compliance field. 
Let $\bm{b}\in\{0,1\}^{H\times W}$ be the drivable-area indicator. 
The outside distance is defined as
\begin{equation}
d_{\mathrm{out}}[i,j]
=
\begin{cases}
0, & \bm{b}[i,j]=1,\\[2mm]
r_{\mathrm{DAC}}
\displaystyle\min_{(u,v):\bm{b}[u,v]=1}
\sqrt{(i-u)^2+(j-v)^2}, & \bm{b}[i,j]=0.
\end{cases}
\end{equation}
The DAC cost field is then
\begin{equation}
    \mathcal{C}_{\mathrm{DAC}}[i,j]
    =
    \max\!\left(0, d_{\mathrm{out}}[i,j]-\epsilon_{\mathrm{safe}}\right),
    \label{eq:dac_cost}
\end{equation}
where $\epsilon_{\mathrm{safe}}$ defines a tolerance band near the drivable-area boundary. 
We isolate the contribution of the field loss in \Cref{tab:train_ablation}.

\paragraph{Total supervised objective.}
The full supervised objective combines masked trajectory generation, structure-aware correction, and drivable-area regularization:
\begin{equation}
    \mathcal{L}_{\mathrm{sup}}
    =
    \mathcal{L}_{\mathrm{DLM}}
    +
    \lambda_{\mathrm{SAP}}\mathcal{L}_{\mathrm{SAP}}
    +
    \lambda_{\mathrm{field}}\mathcal{L}_{\mathrm{field}}.
    \label{eq:sup_loss}
\end{equation}

\subsection{Reinforcement Learning over Draft-and-Edit Rollouts}
\label{sec:method:rl}

Supervised training teaches the model to imitate expert trajectories and recover from synthetic perturbations, but it does not directly optimize closed-loop driving metrics. 
We therefore fine-tune ReflectDrive-2 with reinforcement learning over the composed draft-and-edit rollout. 
The key distinction from a longer single-pass diffusion rollout is that the generation process is explicitly divided into a drafting phase and an AutoEdit phase. 
The terminal reward is assigned to the final post-edit trajectory, and the policy-gradient objective credits token transitions from both phases.

For each scene, we sample $N_g$ goal points by top-$k$ sampling with NMS and draw $I$ drafts per goal, giving $G=N_g I$ candidate rollouts. 
For candidate $g$, the token-transition sequence is
\begin{equation}
    \rho_g =
    \left(
    \xb_g^{0},
    \xb_g^{1},
    \ldots,
    \xb_g^{S_{\mathrm{draft}}},
    \xb_g^{S_{\mathrm{draft}}+1},
    \ldots,
    \xb_g^{S_{\mathrm{draft}}+S_{\mathrm{edit}}}
    \right),
    \label{eq:rollout_sequence}
\end{equation}
where the first $S_{\mathrm{draft}}$ transitions correspond to masked trajectory drafting and the next $S_{\mathrm{edit}}$ transitions correspond to AutoEdit. 
The final trajectory is
\begin{equation}
    \tau_g = \mathrm{Detok}\!\left(\xb_g^{S_{\mathrm{draft}}+S_{\mathrm{edit}}}\right).
\end{equation}
We use the closed-loop planning score as the terminal reward $R(\tau_g)$ and compute a group-relative advantage
\begin{equation}
    A_g
    =
    R(\tau_g)
    -
    \frac{1}{G}\sum_{j=1}^{G}R(\tau_j).
    \label{eq:group_advantage}
\end{equation}

The discrete-diffusion policy-gradient objective in Eq.~\eqref{eq:total_loss} is applied over all token transitions in $\rho_g$. 
Equivalently, the token-transition indicator
\begin{equation}
    \delta_{g,p}^{s}
    =
    \mathbf{1}_{\{x_{g,p}^{s+1}\neq x_{g,p}^{s}\}},
    \qquad
    s=0,\ldots,S_{\mathrm{draft}}+S_{\mathrm{edit}}-1,
    \label{eq:transition_indicator}
\end{equation}
covers both unmasking during drafting and rewriting during AutoEdit. 
The same terminal reward therefore optimizes goal-conditioned drafting and AutoEdit under one rollout objective. 
Because only the post-edit trajectory receives reward, the drafting phase is optimized for trajectories that can be improved by the subsequent correction phase, while AutoEdit is optimized for corrections that improve the closed-loop score rather than only reducing token-level uncertainty.

\section{Efficient Inference for Reflective Masked Planning}
\label{sec:deployment}

We treat deployment as an optimization chain rather than as independent serving tricks.
The optimization sequence is summarized in \Cref{tab:deployment_chain}. 
Each row stacks one optimization on top of the previous and reports the resulting end-to-end planner latency on NVIDIA Thor.
The final stack runs a full decision--draft--reflect pass on full-step frames and a lightweight
temporal AutoEdit pass on lite-step frames, yielding an average latency of $30.2$ ms per frame.

\label{sec:deployment:chain}

\paragraph{Shared-prefix KV reuse.}
Goal-point proposal, trajectory drafting, and AutoEdit all condition on the same visual,
route-instruction, and ego-state prefix. 
Instead of recomputing this prefix for each phase, we keep a shared prefix cache and switch between
\texttt{Single} and \texttt{Batch} cache states according to the current serving phase. 
This reduces the decode latency from $56.8$ ms to $42.2$ ms.

\paragraph{Mutable action-cache rewinding and merged rewrite.}
The action-token block is mutable: masked drafting changes \texttt{[MASK]} tokens into concrete
trajectory tokens, while AutoEdit directly replaces concrete trajectory tokens with revised concrete
tokens. 
In both cases, KV entries associated with the previous action-token state become stale. 
After each update, the cache pointer is rewound to the shared-prefix boundary, and only the mutable
action block is recomputed. 
At multi-block boundaries, we further merge the required cache rewrite with the first token-update
step of the next block, reducing the boundary latency by approximately $3$ ms.
Similarly, by fusing the prefill and keypoint inference into a single pass, the prefill-plus-keypoint stage latency drops from $14.4$ ms to $13.8$ ms.

\paragraph{Action-expert FFN.}
Trajectory-token decoding uses a constrained action vocabulary and a short fixed-length token block.
We therefore replace the full FFN in the action branch with a compact action-expert FFN, reducing the
hidden dimension from $4096$ to $1024$. 
This lowers the decode latency from $42.2$ ms to $29.6$ ms. And we measure the action-expert FFN via trajectory-level metrics summarized in \Cref{tab:deployment_quality_gate} to validate its feasibility. Although minSADE slightly
increases, the compact branch improves meanSADE and the selected path-level error metrics.

\paragraph{Fused on-device token update.}
Both masked drafting and token-to-token AutoEdit require confidence ranking, token selection, and
state update. 
In masked drafting, the update commits predicted tokens in place of \texttt{[MASK]}; in AutoEdit, the
update overwrites selected concrete tokens with replacement tokens. 
A CPU implementation introduces device synchronization at every step. 
We fuse token selection, ranking, and token-state update into an on-device CUDA kernel, reducing
the per-step update latency from $29.6$ ms to $27.3$ ms.

\begin{table}[t]
  \centering
  \caption{\textbf{Inference optimization chain on NVIDIA Thor.}
  All latencies in ms. Each row stacks one optimization on top of the previous;
  ``---'' denotes unchanged from the row above.
  On ASD lite-step frames, keypoint proposal(Decision) is skipped and only two decode steps are run.}
  \label{tab:deployment_chain}
  \small
  \setlength{\tabcolsep}{6pt}
  \renewcommand{\arraystretch}{1.2}
  \begin{tabular}{@{}>{\raggedright\arraybackslash}p{0.36\linewidth}
                    >{\centering\arraybackslash}p{0.21\linewidth}
                    >{\centering\arraybackslash}p{0.15\linewidth}
                    >{\centering\arraybackslash}p{0.13\linewidth}@{}}
    \toprule
    \textbf{Optimization} &
    \textbf{Prefill + Decision} &
    \textbf{Decode} &
    \textbf{Total} \\
    \midrule
    Baseline                    & 14.4 & 56.8 & 71.2 \\
    +\,Merged infer           & 13.8 & ---  & 70.6 \\
    +\,Shared-prefix KV reuse   & ---  & 42.2 & 56.0 \\
    +\,Action-expert FFN        & ---  & 29.6 & 43.4 \\
    +\,Fused CUDA unmasking     & ---  & 27.3 & \textbf{41.1}\,\textcolor{muted}{\scriptsize(full-step)} \\
    +\,ASD (lite-step) & 9.8 & 9.5 & \textbf{19.3}\,\textcolor{muted}{\scriptsize(lite-step)} \\
    \bottomrule
  \end{tabular}
\end{table}

\begin{table}[t]
  \centering
  \caption{\textbf{Quality gates for deployment optimizations.}
  $\Delta$ denotes optimized $-$ baseline. \dgood{Green} marks the desired direction; \dbad{red} marks regressions. For driving scores ($\uparrow$) higher is better; for trajectory errors ($\downarrow$) lower is better.}
  \label{tab:deployment_quality_gate}
  \small
  \setlength{\tabcolsep}{4pt}
  \renewcommand{\arraystretch}{1.15}

  \begin{tabular}{@{}lccccccc@{}}
    \toprule
    \multicolumn{8}{@{}l}{\textit{(a) ASD temporal AutoEdit on the in-house deployment benchmark} ($\uparrow$)} \\
    \midrule
    \textbf{Setting}            & \textbf{NC} & \textbf{DAC} & \textbf{TTC} & \textbf{Comf.} & \textbf{EP} & \textbf{Overall} & \\
    \midrule
    Full pipeline (baseline)    & 79.51 & 76.99 & 96.27 & 88.52 & 75.13 & 83.70 & \\
    ASD                         & 79.19 & 77.21 & 96.20 & 87.88 & 75.01 & 83.51 & \\
    \midrule
    $\Delta$                    & \dbad{$-0.31$} & \dgood{$+0.22$} & \dbad{$-0.07$} & \dbad{$-0.63$} & \dbad{$-0.12$} & \dbad{$-0.20$} & \\
    \bottomrule
  \end{tabular}

  \vspace{0.7em}

  \begin{tabular}{@{}lcc cccccc@{}}
    \toprule
    \multicolumn{9}{@{}l}{\textit{(b) Action-expert FFN trajectory quality} ($\downarrow$)} \\
    \midrule
    & \multicolumn{2}{c}{\textbf{Trajectory-level}}
    & \multicolumn{3}{c}{\textbf{path-meanFDE}}
    & \multicolumn{3}{c}{\textbf{path-minADE}} \\
    \cmidrule(lr){2-3} \cmidrule(lr){4-6} \cmidrule(lr){7-9}
    \textbf{Setting}
      & \textbf{minSADE} & \textbf{meanSADE}
      & \textbf{@20} & \textbf{@40} & \textbf{@80}
      & \textbf{@20} & \textbf{@40} & \textbf{@80} \\
    \midrule
    Original FFN (baseline)
      & 0.858 & 1.710
      & 0.167 & 0.427 & 0.991
      & 0.088 & 0.210 & 0.462 \\
    Action-expert FFN
      & 0.914 & 1.539
      & 0.145 & 0.346 & 0.910
      & 0.084 & 0.177 & 0.406 \\
    \midrule
    $\Delta$
      & \dbad{$+0.056$} & \dgood{$-0.171$}
      & \dgood{$-0.022$} & \dgood{$-0.081$} & \dgood{$-0.081$}
      & \dgood{$-0.004$} & \dgood{$-0.033$} & \dgood{$-0.056$} \\
    \bottomrule
  \end{tabular}
\end{table}
\paragraph{Alternating Step Decode as temporal token-to-token AutoEdit.}
In streaming driving, adjacent frames share scene context and future plans. 
ReflectDrive-2 therefore alternates between full-step and lite-step frames. 
A full-step frame runs the complete decision--draft--reflect pipeline. 
A lite-step frame transforms the previous plan into the current ego frame and applies a short
token-to-token AutoEdit update instead of rebuilding the trajectory from scratch:
\begin{equation}
  \tilde{\tau}^{(t)}
  =
  \mathcal{T}_{t-1\rightarrow t}
  \!\left(
    \operatorname{Shift}\big(\hat{\tau}^{(t-1)}\big)
  \right),
  \qquad
  \xb^{(t,0)}
  =
  \operatorname{Tok}\!\left(\tilde{\tau}^{(t)}\right),
  \qquad
  \hat{\xb}^{(t)}
  =
  \operatorname{T2TEdit}^{S'}_{\theta}
  \!\left(
    \xb^{(t,0)}, \mathbf{c}^{(t)}
  \right).
  \label{eq:asd_t2t}
\end{equation}
Here $\operatorname{Shift}(\cdot)$ removes the elapsed portion of the previous plan, and
$\mathcal{T}_{t-1\rightarrow t}$ transforms the remaining waypoints from the previous ego frame to
the current ego frame. 
We use $S={1+3+3}$ decision--draft--reflect steps for full-step frames and $S'={1+1}$ draft--reflect steps for lite-step frames, reducing the decode latency from $27.3$ ms to $9.5$ ms. And we also evaluate ASD against running the full pipeline on
every frame in \Cref{tab:deployment_quality_gate}. Replacing alternating full frames with temporal token-to-token AutoEdit changes the
in-house overall score by only $-0.20$, while drivable area compliance slightly improves. Overall, the resulting
planner runs at 30.2 ms average latency on NVIDIA Thor, with full-step frames at 41.1 ms and
lite-step frames at 19.3 ms. Thus, the same token-to-token AutoEdit operator serves both as an intra-frame trajectory corrector and
as an inter-frame temporal refiner.
\section{Experiments}
\label{sec:experiments}

\subsection{Experimental Setup}
\label{sec:exp:setup}

\paragraph{Dataset and metrics.}
We evaluate ReflectDrive-2 on NAVSIM~\citep{dauner2024navsim}, a closed-loop planning benchmark built on nuPlan~\citep{caesar2021nuplan}. The task is to predict a 4-second ego trajectory at 2 Hz. We train on navtrain ($1{,}192$ scenes) and evaluate on navtest ($136$ scenes). The metric is Predictive Driver Model Score (PDMS), aggregating no at-fault collision (NC), drivable-area compliance (DAC), time to collision (TTC), comfort, and ego progress (EP).

\paragraph{Implementation.}
A $0.7$B masked-diffusion language backbone and a $0.1$B ViT visual encoder, both initialized from proprietary pretrained weights, are fully fine-tuned on NAVSIM. The input is two temporal frames from the left-front/front/right-front cameras plus navigation instruction and ego-state tokens; the output is $8$ waypoints represented as $16$ discrete coordinate tokens. We supervise-fine-tune first, then reinforcement-fine-tune with PDMS as reward. 

\paragraph{Baselines.}
End-to-end planners: UniAD~\citep{hu2023uniad}, TransFuser~\citep{chitta2022transfuser}, Hydra-MDP~\citep{zhang2024hydramdp}, DiffusionDrive~\citep{li2024diffusiondrive}, GoalFlow~\citep{zhou2024goalflow}. VLA planners: AutoVLA~\citep{zhou2024autovla}, DriveVLA-W0~\citep{jin2024drivevla}, ReCogDrive~\citep{zhang2024recogdrive}. For standard evaluation all methods emit one trajectory; for best-of-$N$ evaluation, ReflectDrive-2 samples multiple goal points and keeps the trajectory with the highest closed-loop score (oracle selection). In standard evaluation, we use the highest-confidence goal after NMS and output one trajectory. Best-of-6 evaluation reports the oracle-best over six candidate trajectories sampled from six goal-point proposals; during RFT the same group size of six is composed as three goal points with two drafts each.

\subsection{Effect of RL on Inference-Time AutoEdit}
\label{sec:exp:rl_autoedit}

\begin{table}[t]
  \centering
  \caption{\textbf{Effect of inference-time AutoEdit across training regimes.} After supervised training, inference-time AutoEdit contributes at most $+0.3$ PDMS. After RL over the full draft-and-edit rollout, the same inference-time AutoEdit contributes $+1.9$ PDMS.}
  \label{tab:autoedit_gain}
  \small
  \setlength{\tabcolsep}{7pt}
  \begin{tabular}{lccc}
    \toprule
    \textbf{Training setting} & \textbf{w/o AutoEdit} & \textbf{w/ AutoEdit} & \textbf{$\Delta$ PDMS} \\
    \midrule
    DLM & 84.8 & 85.0 & +0.2 \\
    DLM + DACF & 87.2 & 87.3 & +0.1 \\
    DLM + DACF + AutoEdit training & 87.7 & 88.0 & +0.3 \\
    \rowcolor{rowE2E}
    DLM + DACF + AutoEdit training + RL & 89.1 & \textbf{91.0} & \textbf{+1.9} \\
    \bottomrule
  \end{tabular}
\end{table}

\Cref{tab:autoedit_gain} isolates the main interaction behind our final result. Before RL, inference-time AutoEdit delivers at most $+0.3$ PDMS, regardless of whether AutoEdit was trained with structure-aware perturbation -- the editor is learned, but its closed-loop contribution remains modest. After RL over the full draft-and-edit rollout, the same inference-time AutoEdit contributes $+1.9$ PDMS, a substantial increase relative to the supervised AutoEdit gain. The mechanism is the interaction described in \Cref{sec:method:rl}: with a shared terminal reward, the drafter learns to emit revisable drafts (token distributions whose post-edit score exceeds their pre-edit score), and AutoEdit learns corrections that move the draft toward reward rather than only reducing token-level uncertainty. This interaction requires a composed draft-and-edit rollout and does not arise in single-pass RL formulations (\Cref{sec:related:rl}). The $89.1\!\to\!91.0$ improvement that distinguishes ReflectDrive-2 from the strongest camera-only VLA on NAVSIM (\Cref{tab:navsim}) comes from this interaction.

\subsection{Closed-Loop Driving Performance}
\label{sec:exp:closedloop}

\begin{table}[t]
  \centering
  \caption{\textbf{Closed-loop planning results on NAVSIM.} All methods evaluated under standard single-trajectory setting. ``C \& L'' denotes camera and LiDAR.}
  \label{tab:navsim}
  \small
  \setlength{\tabcolsep}{4pt}
  \begin{tabular}{llcccccc}
    \toprule
    \textbf{Method} & \textbf{Input} & \textbf{NC$\uparrow$} & \textbf{DAC$\uparrow$} & \textbf{TTC$\uparrow$} & \textbf{Comf.$\uparrow$} & \textbf{EP$\uparrow$} & \textbf{PDMS$\uparrow$} \\
    \midrule
    \multicolumn{8}{l}{\textit{End-to-End Methods}} \\
    UniAD~\citep{hu2023uniad} & Cam & 97.8 & 91.9 & 92.9 & 100.0 & 78.8 & 83.4 \\
    TransFuser~\citep{chitta2022transfuser} & C \& L & 97.7 & 92.8 & 92.8 & 100.0 & 79.2 & 84.0 \\
    Hydra-MDP~\citep{zhang2024hydramdp} & C \& L & 98.3 & 96.0 & 94.6 & 100.0 & 78.7 & 86.5 \\
    DiffusionDrive~\citep{li2024diffusiondrive} & C \& L & 98.2 & 96.2 & 94.7 & 100.0 & 82.2 & 88.1 \\
    GoalFlow~\citep{zhou2024goalflow} & C \& L & 98.4 & 98.3 & 94.6 & 100.0 & 85.0 & 90.3 \\
    \midrule
    \multicolumn{8}{l}{\textit{Camera-Only VLA Planners}} \\
    AutoVLA~\citep{zhou2024autovla} & Cam & 98.4 & 95.6 & 98.0 & 99.9 & 81.9 & 89.1 \\
    DriveVLA-W0~\citep{jin2024drivevla} & Cam & 98.7 & 99.1 & 95.3 & 99.3 & 83.3 & 90.2 \\
    ReCogDrive~\citep{zhang2024recogdrive} & Cam & 97.9 & 97.3 & 94.9 & 100.0 & 87.3 & 90.8 \\
    \textbf{ReflectDrive-2 (Ours)} & Cam & 97.3 & 98.1 & 92.5 & 100.0 & \textbf{89.4} & \textbf{91.0} \\
    \bottomrule
  \end{tabular}
\end{table}

\Cref{tab:navsim} presents standard single-trajectory results as downstream evidence that the RL$\times$AutoEdit coupling works end-to-end. ReflectDrive-2 reaches $91.0$ PDMS with camera-only input, above ReCogDrive ($90.8$, camera-only) and GoalFlow ($90.3$, camera and LiDAR). The largest gain is in ego progress (EP $=89.4$, highest among listed methods), while DAC remains high at $98.1$ and comfort is saturated at $100.0$. The result suggests a favorable progress--constraint trade-off: EP improves substantially while DAC and comfort remain high, although NC and TTC are not the best among the listed baselines. Camera-only is the harder setting, and the comparison of interest is against the other camera-only VLA peers (AutoVLA / DriveVLA-W0 / ReCogDrive), over which ReflectDrive-2's $+0.2$ to $+1.9$ PDMS advantage is driven by the rollout-level RL interaction isolated in \Cref{tab:autoedit_gain}.

\begin{table}[t]
  \centering
  \caption{\textbf{Best-of-$N$ evaluation on NAVSIM.} Best-of-6 uses oracle PDMS selection and is reported to measure the quality of the goal-point posterior, not as a standard benchmark result.}
  \label{tab:bestofn}
  \small
  \setlength{\tabcolsep}{6pt}
  \begin{tabular}{lcccccc}
    \toprule
    \textbf{Setting} & \textbf{NC$\uparrow$} & \textbf{DAC$\uparrow$} & \textbf{TTC$\uparrow$} & \textbf{Comf.$\uparrow$} & \textbf{EP$\uparrow$} & \textbf{PDMS$\uparrow$} \\
    \midrule
    ReflectDrive-2 (single) & 97.3 & 98.1 & 92.5 & 100.0 & 89.4 & 91.0 \\
    ReflectDrive-2 (best-of-6, oracle) & 98.5 & 99.2 & 95.5 & 99.8 & \textbf{93.8} & \textbf{94.8} \\
    Human (NAVSIM reference) & 100.0 & 100.0 & 100.0 & 99.9 & 87.5 & \textbf{94.8} \\
    \bottomrule
  \end{tabular}
\end{table}

\Cref{tab:bestofn} shows that best-of-$6$ ReflectDrive-2 reaches $94.8$ PDMS, reaching the NAVSIM human reference under oracle selection. The gap between single and best-of-$6$ measures the quality of the goal-point posterior: $3.8$ PDMS of headroom is recovered by selecting a different behavior hypothesis per scene, evidence that the goal head exposes a genuinely multi-modal action posterior rather than noisy replicas of the same endpoint.

\subsection{Decision Diversity and Reflection}
\label{sec:exp:decision_reflection}

\begin{figure}[h]
  \centering
  \includegraphics[width=\linewidth]{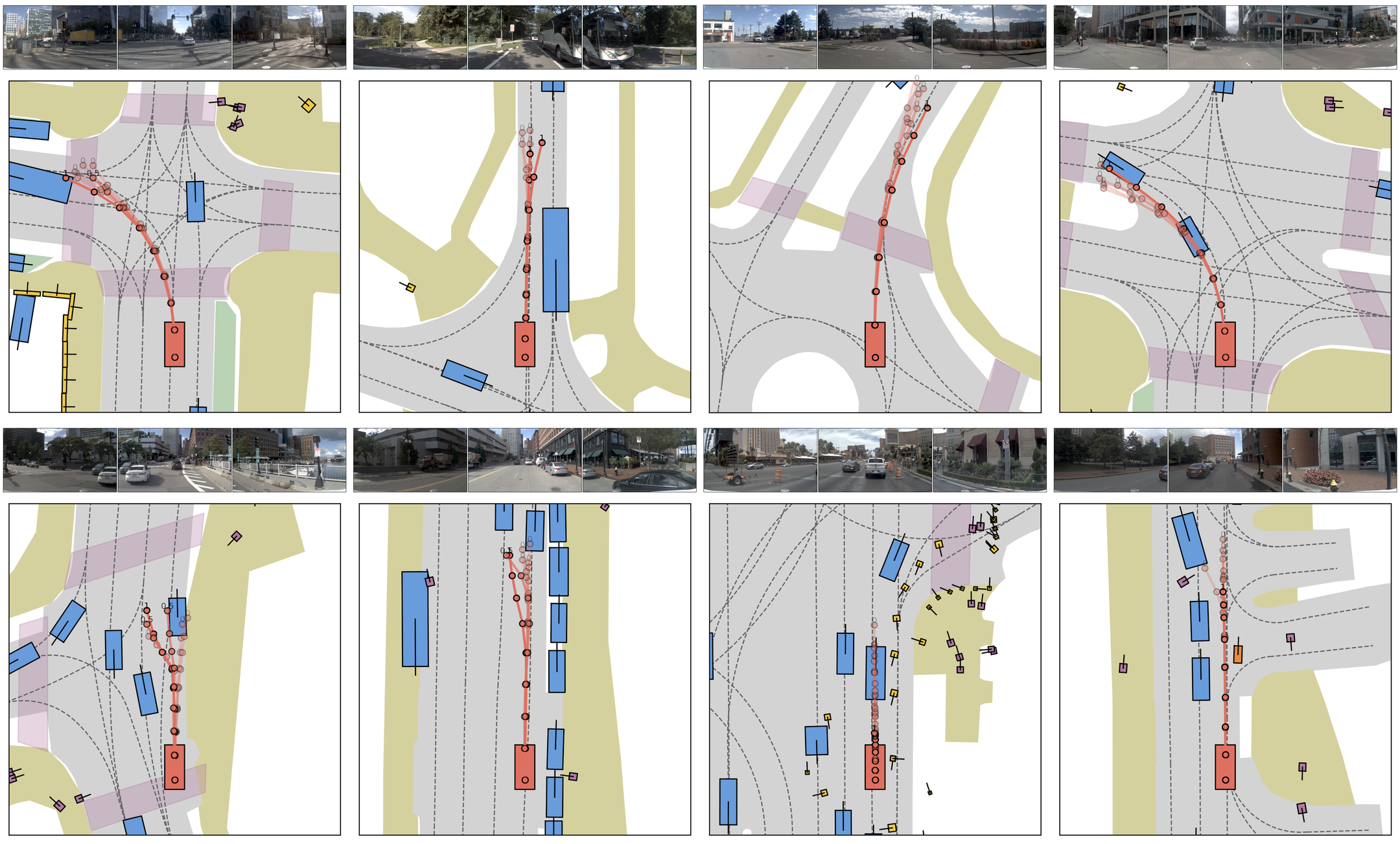}
  \caption{\textbf{Decision diversity from goal points.} Each goal anchors a distinct behavior hypothesis. Trajectory opacity encodes PDMS.}
  \label{fig:gp_qualitative}
\end{figure}

\paragraph{Goal points.}
\Cref{fig:gp_qualitative} visualizes multi-goal inference. In turning scenes (top row), different goal points realize different lines through the curve and some candidates respect the drivable boundary better than others. In interaction scenes (bottom row), the model produces longitudinally and laterally distinct behaviors around nearby agents (keep lane / change lane / adjust speed). Goal points are not sampling noise; they are distinct behavior hypotheses that the downstream decoder realizes.

\begin{figure}[t]
  \centering
  \includegraphics[width=\linewidth]{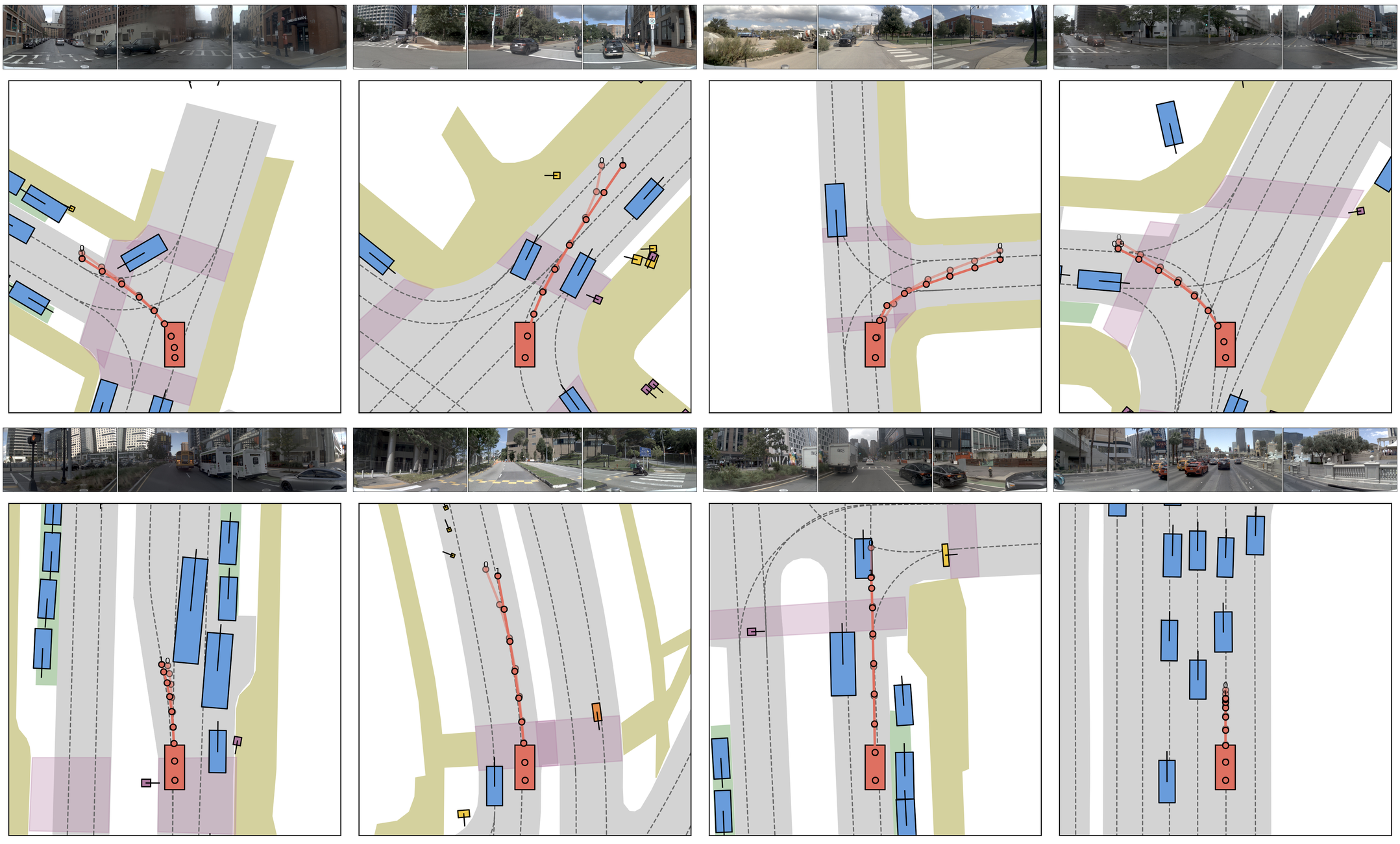}
  \caption{\textbf{Inference-time reflection with AutoEdit.} Semi-transparent: initial drafts. Solid: final outputs after AutoEdit. PDMS annotated at trajectory endpoints.}
  \label{fig:autoedit_qualitative}
\end{figure}
\paragraph{Reflection with AutoEdit.}
\Cref{fig:autoedit_qualitative} shows AutoEdit at inference. Semi-transparent curves are initial drafts; solid curves are post-AutoEdit. In the top row, AutoEdit pulls trajectories back into the drivable area; in the bottom row, it adjusts the plan around nearby agents. The revisions are structured rewrites in the same token space used for generation, not cosmetic smoothing.

\begin{figure}[h]
  \centering
  \begin{minipage}[t]{0.48\linewidth}
    \centering
    \includegraphics[width=\linewidth]{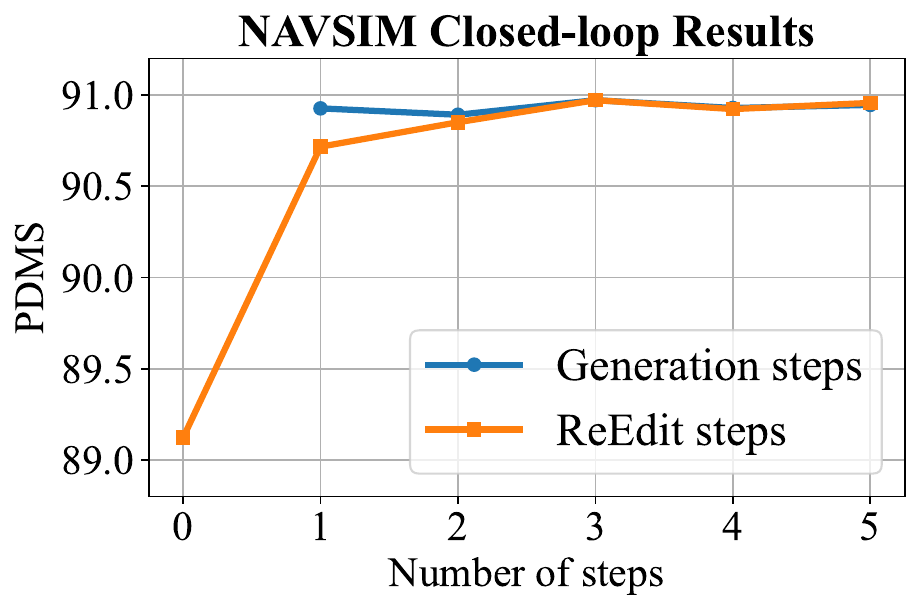}
    \caption{\textbf{Sensitivity to diffusion steps.} PDMS with different generation and AutoEdit step counts.}
    \label{fig:param_ablation}
  \end{minipage}
  \hfill
  \begin{minipage}[t]{0.47\linewidth}
    \centering
    \includegraphics[width=\linewidth]{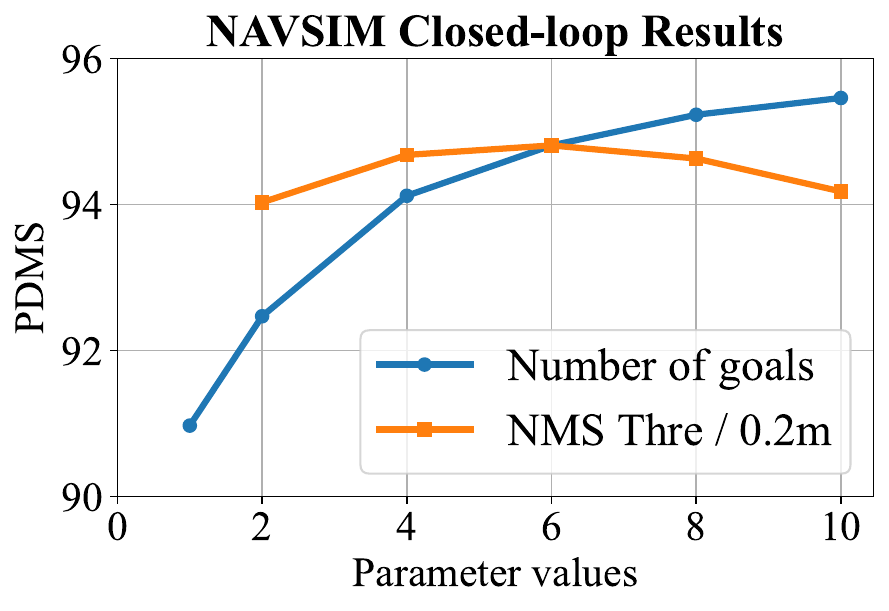}
    \caption{\textbf{Sensitivity to goal-proposal parameters.} PDMS vs.\ number of proposals and NMS threshold.}
    \label{fig:bestofn_ablation}
  \end{minipage}
\end{figure}
\subsection{Ablation Studies}
\label{sec:exp:ablation}

\paragraph{Training components.}
\Cref{tab:train_ablation} isolates the effect of each training component without inference-time AutoEdit. Field loss (DACF) contributes $+2.4$ PDMS ($84.8\!\to\!87.2$) mainly through DAC. AutoEdit supervised training adds $+0.5$ more. RL over the full rollout brings EP from $82.2$ to $89.3$ and PDMS to $89.1$; combined with inference-time AutoEdit, the final score is $91.0$ (\Cref{tab:autoedit_gain}).

\begin{table}[t]
  \centering
  \caption{\textbf{Effect of training components.} All rows are evaluated without inference-time AutoEdit to isolate the training signal.}
  \label{tab:train_ablation}
  \small
  \setlength{\tabcolsep}{6pt}
  \begin{tabular}{lcccccc}
    \toprule
    \textbf{Training objective} & \textbf{NC$\uparrow$} & \textbf{DAC$\uparrow$} & \textbf{TTC$\uparrow$} & \textbf{Comf.$\uparrow$} & \textbf{EP$\uparrow$} & \textbf{PDMS$\uparrow$} \\
    \midrule
    DLM & 97.5 & 93.9 & 92.8 & 99.5 & 79.5 & 84.8 \\
    DLM + DACF & 97.4 & 97.0 & 93.1 & 99.9 & 81.4 & 87.2 \\
    DLM + DACF + AutoEdit training & \textbf{97.8} & 96.7 & \textbf{93.6} & 99.9 & 82.2 & 87.7 \\
    DLM + DACF + AutoEdit training + RL & 96.3 & \textbf{97.9} & 88.9 & 99.7 & \textbf{89.3} & \textbf{89.1} \\
    \bottomrule
  \end{tabular}
\end{table}

\paragraph{Inference budget.}
\Cref{fig:param_ablation} sweeps generation steps and AutoEdit steps: performance improves and then plateaus around $3$--$5$ steps, consistent with masked diffusion -- a small number of rounds forms a coherent trajectory, while excess rewriting disturbs a good draft.

\paragraph{Goal-proposal parameters.}
\Cref{fig:bestofn_ablation} varies the number of goal proposals and the NMS threshold in best-of-$N$. More proposals expose more behavior hypotheses; an NMS threshold near $1.2$ m is optimal -- smaller keeps duplicates, larger removes genuine alternatives.

\section{Conclusion}
\label{sec:conclusion}

We presented \textit{ReflectDrive-2}, a reflective VLA planner that reframes autonomous driving as a joint process of decision making, trajectory drafting, and self-correction, all within a shared discrete token space. A goal-point posterior exposes behavior-level hypotheses before low-level motion generation, masked discrete diffusion drafts editable trajectories in parallel, and AutoEdit rewrites the draft through the same policy without an auxiliary repair network.
The central finding of this work is that self-correction in driving planning requires more than a trained editor. Under supervised training alone, AutoEdit exists in the weights but contributes only modestly at inference. Applying reinforcement learning over the complete draft-and-edit rollout changes this: a shared terminal reward co-adapts the drafter and the editor, so that drafts become revisable and edits become reward-seeking. This interaction raises the inference-time AutoEdit gain from $+0.3$ to $+1.9$ PDMS and is the primary driver of ReflectDrive-2's $91.0$ PDMS on NAVSIM with camera-only input. Under best-of-$6$ oracle selection, the system reaches $94.8$ PDMS, indicating that the goal-point posterior captures a genuinely multi-modal distribution over driving behaviors.
We further showed that the decision--draft--reflect structure defines not only a modeling paradigm but also an efficient runtime. Shared-prefix KV cache reuse, Alternating Step Decode, a lightweight action-expert FFN, and fused on-device unmasking together bring the average planner-stack latency to $30.2$ ms per frame on NVIDIA Thor, with near-lossless planning quality. These results suggest that masked discrete diffusion can serve as an editable and deployable foundation for VLA driving policies.


\paragraph{Limitations and future work.}
ReflectDrive-2 represents trajectories with fixed-resolution BEV coordinate tokens. This choice provides an interpretable and editable action space for masked drafting and AutoEdit, but it also bounds the spatial precision of the generated waypoints by the coordinate-bin size. Future work could improve precision with finer coordinate vocabularies, residual offsets, or hybrid discrete-continuous action heads while retaining token-space editability. Our RL stage currently optimizes a lightweight closed-loop planning score, which is efficient for post-training but remains a proxy for real-world driving objectives. Higher-fidelity interactive simulators and richer safety-oriented rewards may improve alignment, albeit with higher computational cost. In addition, the current AutoEdit perturbations focus on longitudinal progress and lateral heading errors; extending them to interaction-level failures such as yielding timing, cut-in response, and gap selection could improve correction in multi-agent scenes.

\bibliographystyle{plainnat}
\bibliography{references}

\end{document}